\documentclass[letterpaper, 10 pt, conference]{ieeeconf}  %

\usepackage{comment}
\usepackage{cite}
\usepackage[ruled,vlined]{algorithm2e}
\usepackage{stfloats}
\usepackage{amsmath,amssymb,amsfonts}
\usepackage{algpseudocode} %
\usepackage{graphicx}
\usepackage{textcomp}
\usepackage{xcolor}
\usepackage{booktabs}
\usepackage{lipsum}
\usepackage{balance}
\usepackage[dvipsnames]{xcolor}
\def\BibTeX{{\rm B\kern-.05em{\sc i\kern-.025em b}\kern-.08em
    T\kern-.1667em\lower.7ex\hbox{E}\kern-.125emX}}
\IEEEoverridecommandlockouts                              %

\title{\LARGE \bf
Contact-Driven Localization in a \\Freeform Robotic Self-Assembled Structure
}

\author{Mohammadali Rashidioun$^{1}$, Michael Sosa$^{2,*}$, Petras Swissler$^{3}$
\thanks{$^{1}$ Industrial and Mechanical Engineering Department, 
        New Jersey Institute of Technology. 
        {\tt\small Mr943@njit.edu}}
\thanks{$^{2}$ Ying Wu College of Computing, New Jersey Institute of Technology. {\tt\small ms3383@njit.edu}}
\thanks{$^{3}$ Industrial and Mechanical Engineering Department, 
        New Jersey Institute of Technology.
        {\tt\small Petras.Swissler@njit.edu}}
        \thanks{$^{*}$ Work done prior to current employment}
}

\usepackage{tikz}

\newcommand\copyrighttext{%
  \footnotesize © 2026 IEEE. Personal use of this material is permitted. Permission from IEEE must be obtained for all other uses, in any current or future media, including reprinting/republishing this material for advertising or promotional purposes, creating new collective works, for resale or redistribution to servers or lists, or reuse of any copyrighted component of this work in other works.}

\newcommand\copyrightnotice{%
  \begin{tikzpicture}[remember picture,overlay]
    \node[anchor=south,yshift=10pt] at (current page.south) {\fbox{\parbox{\dimexpr\textwidth-\fboxsep-\fboxrule\relax}{\copyrighttext}}};
  \end{tikzpicture}%
}

\begin{document}

\maketitle

\copyrightnotice

\thispagestyle{empty}
\pagestyle{empty}

\begin{abstract}Accurate localization remains a key challenge in swarm robotics, particularly for self-reconfigurable systems that must identify relative positions to form diverse structures. Most existing approaches rely on external tracking infrastructure or high-cost sensors, which limit scalability and deployment in unstructured environments. In this paper, we propose a novel contact-driven localization method for modular robots that leverages only local communication through binary contact information (whether two robots are physically connected or not). To exploit these contact cues, we introduce a virtual-force framework in which robots iteratively refine their poses—attracting toward dock-connected neighbors and repelling from non-connected ones. The method requires no external infrastructure and relies only on minimal onboard sensing. Simulations show effective localization during the assembly of towers and cantilevers, enabling accurate, scalable, free-form self-assembly.

\end{abstract}

\section{INTRODUCTION}
\subsection{Background}
    
A fundamental requirement for coordinated multi-robot behavior is \emph{localization}~\cite{wang_distributed_2023}. 
Swarm robotics—multi-robot systems inspired by the collective behaviors of social animals—uses large groups of simple robots that cooperate to perform tasks beyond the capability of a single agent~\cite{brambilla_swarm_2013,TAN201318}. 
In such swarms, agents must estimate their relative positions to maintain formations, avoid collisions, and achieve precise objectives in applications such as environmental monitoring~\cite{KHALDI2025100365}, disaster response~\cite{10869410}, and autonomous self-assembly—even in environments where external infrastructure is unavailable~\cite{s22124424}.

Absolute positioning using Global Navigation Satellite System (GPS, BeiDou, GLONASS, Galileo) \cite{rs14163930} is mostly impractical in densely packed swarms or obstructed environments (underground, mountainous terrain) \cite{s23052399}. Accordingly, alternative localization strategies, each with their own set of limitations, have been proposed across different platforms to address these challenges. These strategies often fall into two categories, either being centralized through an external controller or observer or decentralized within the swarm.
Several swarm platforms achieve localization by relying on overhead or environment-mounted systems. 
Coachbots~\cite{8463206} used fiducial tags mounted along arena walls, with cameras onboard estimating pose relative to detected tags. 
Later, in its second version~\cite{9000788}, VR-based localization using infrared timing signals replaced the fiducial tags, which were decoded by onboard receivers for pose estimation. 
Zooids~\cite{le2016zooids} employed projector-based optical patterns read by photodiodes on each robot, while SwarmBang~\cite{duan2023minimalist} tracked ceiling beacons using upward-facing cameras. 
All of the mentioned methods are dependent on external installations and are largely confined to flat arenas, and thus cannot be easily generalized to unknown or outdoor environments.

\begin{figure}
  \centering
  \includegraphics[width=1\linewidth]{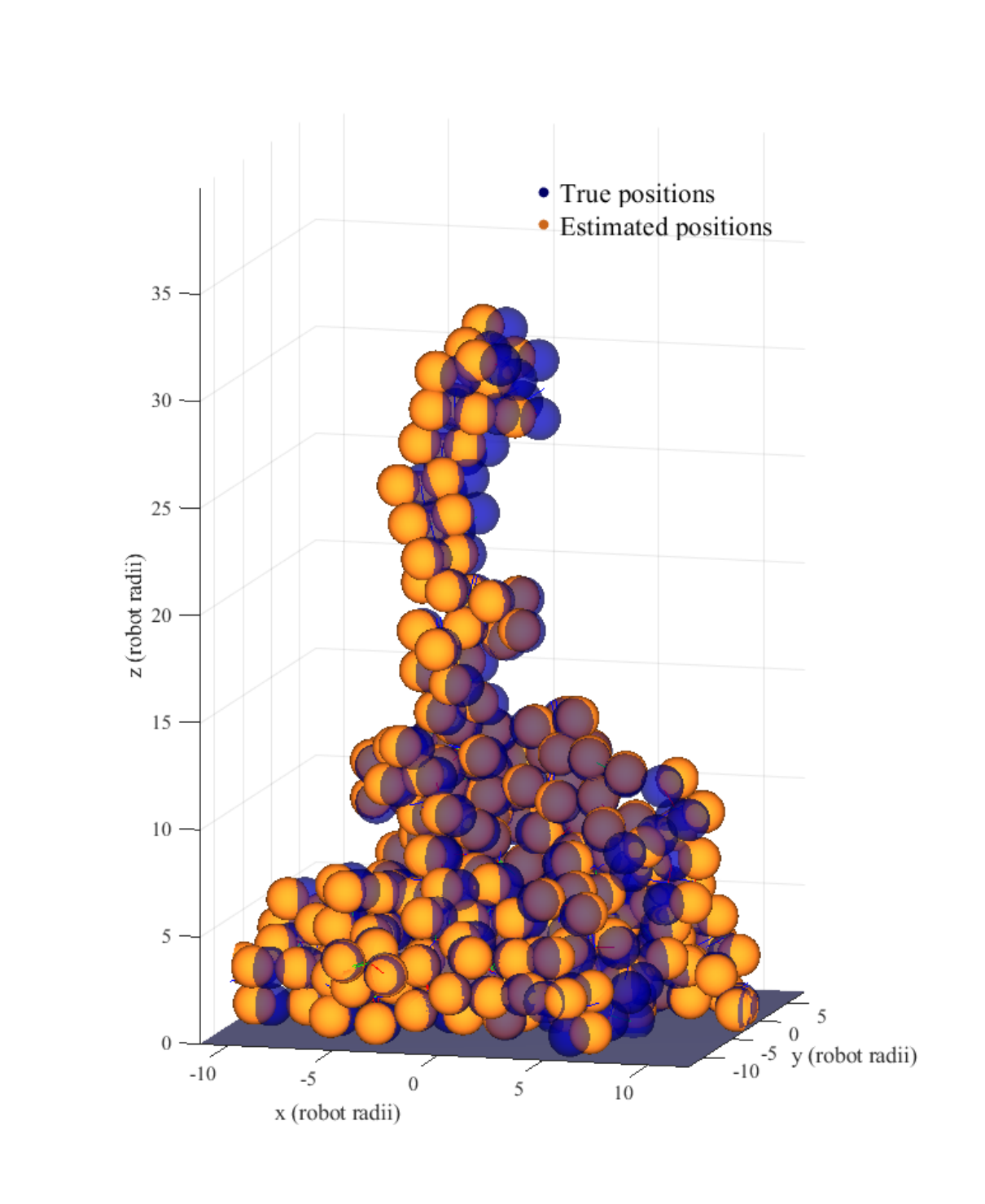}
  \caption{Contact-driven localization in a ReactiveBuild-assembled tower. Blue spheres indicate ground-truth positions; orange spheres indicate estimated positions.}
  \label{fig:hero}
\end{figure}

In contrast to methods that use external installations, decentralized methods rely only on onboard sensing.
Among decentralized 2D approaches, Kilobots\cite{rubenstein2014programmable} use range-only infrared sensing `seed'' robots to initialize a shared frame that then propagates hop-by-hop via distance gradients. This removes the need for external observers, but remains planar; errors accumulate with hop count, and the absence of bearing measurements limits orientation accuracy.

Bridging to 3D, aerial swarms routinely achieve precise 3D formations under external motion-capture (e.g., Vicon), but that centralizes estimation and ties deployment to lab infrastructure (not onboard sensing)~\cite{kushleyev2012towards}.

Some modular platforms adopt lattice architectures. M-Blocks \cite{8967810} use magnetic fiducial tags and absolute magnetic encoders to identify neighbors and connection orientation \cite{8967810}. Although compact and passive, accuracy depends on precise face alignment and manufacturing tolerances. Because sensing requires face-to-face alignment, localization is naturally tied to axis-aligned lattice configurations.

FreeSN enables 3D freeform self-assembly through configuration identification, which combines dense magnetic sensing, inertial measurements, and odometry. A central computer aggregates these outputs to reconstruct the topology \cite{tu2022freesn}. Despite its capability, this approach is hardware-intensive and depends on centralized processing.

Table~\ref{tab:localization_cost} summarizes the localization methods discussed above.
    As you can see, most of the centralized 2D approaches rely on external infrastructure and line-of-sight; decentralized 2D approaches suffer error accumulation and do not readily extend to 3D; and freeform 3D systems offer expressiveness at substantial hardware and computational cost.  %

\begin{table*}[t]
\centering
\caption{Sensor cost per module and main limitations.}
\resizebox{\linewidth}{!}{%
\begin{tabular}{|l|c|p{8cm}|}
\hline
\textbf{System} & \textbf{Approximate Cost} & \textbf{Description} \\
\hline
Coachbot V1 \cite{espinosa2018using}& \$20 & 2D localization using a camera and external ArUco tags. \\
\hline
Coachbot \cite{dornadula2024Coachbot} & \$20 sensor + \$150 base & 2D localization using IR signals via VR lighthouse.\\
\hline
Zooids \cite{le2016zooids} &\$1 sensor + \$1000 projector & High-precision 2D localization using high-speed structured light.\\
\hline
Kilobot \cite{rubenstein2014programmable} &  Negligible& Decentralized 2D localization via range sensing and communication. \\
\hline
Free-SN \cite{tu2022freesn} & \$150 & Freeform 3D localization via 42 magnetometers + IMU; centralized. \\
\hline
M-Blocks \cite{8967810} & \$50& 3D Lattice-based localization using face tags.\\
\hline
ATRON \cite{brandt2007atron} & Negligible & 3D Lattice-based localization using IR communication\\
\hline
SwarmBang \cite{duan2023minimalist} & \$50 & 2D localization using line-of-sight to moving beacon\\
\hline
Our method & \$20 for FireAntV3 \cite{swissler2023fireantv3}  & Freeform 3D contact-driven localization.\\
\hline
\end{tabular}
}
\label{tab:localization_cost}
\end{table*}

This paper presents an approach to robot localization in a 3D freeform self-assembled structure using modest sensing requirements.
The major contributions of the presented work are as follows:
\begin{enumerate}
    \item A method to enable freeform modular self-assembling robots to localize within a self-assembled structure.
    \item Insight regarding the importance of robot physical characteristics in the context of localization.
    \item Insight into sensing requirements for freeform self-assembling robotic systems.
\end{enumerate}

The remainder of the paper is organized as follows: First, we present the unique characteristics of a FireAnt robot that provide additional localization information inherent in the physical design. Next, we present the methodology for using robot contacts to localize robots within a self-assembled structure. Finally, we examine several cases for sensor availability and present the localization quality for each.
\section{FireAnt Design Features for Localization}
\begin{figure*}[!t]
    \centering
    \includegraphics[width=\textwidth, height=8cm, keepaspectratio]{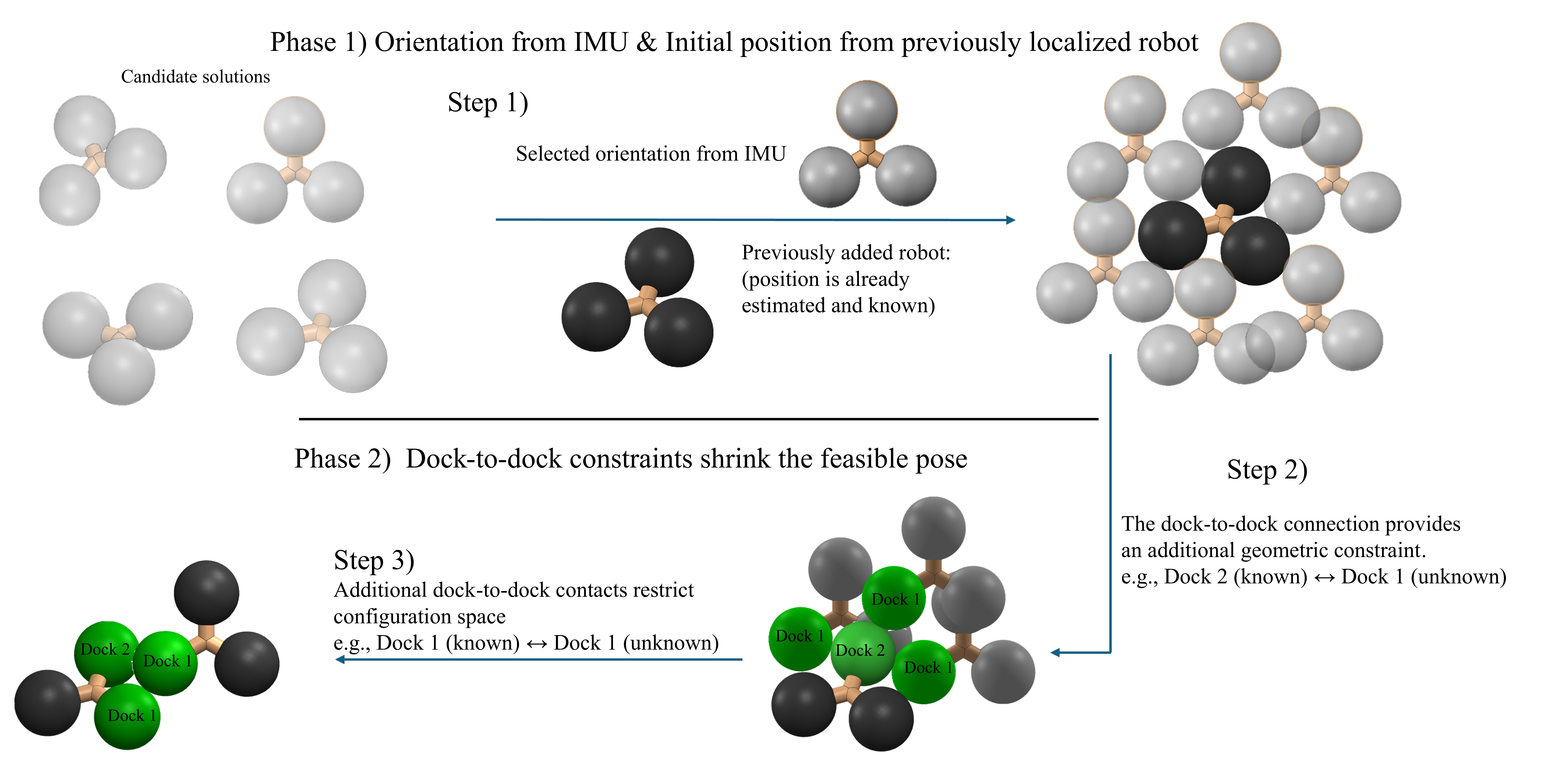} 
    \caption[Contact-driven localization constraints with FireAntV3]{
    \textbf{Contact-driven localization constraints with FireAntV3.}
    \textbf{Phase 1 (top):} Proprioceptive sensing (IMU) constrains the unknown robot’s orientation to a finite set of candidate poses. In addition, its connection to a previously placed robot provides an initial positional reference that further restricts feasible locations. 
    \textbf{Phase 2 (bottom):} Dock-to-dock identities impose additional geometric constraints. For example, if Dock~2 of the known robot connects to Dock~1 of the unknown robot, the relative position is bounded by the fixed geometry of tangential spheres.  Each additional dock-to-dock contact further constrains the pose, yielding a near-unique localization.}
    \label{fig:contact_constraints}
\end{figure*}

In most modular systems, localization is minimally constrained and high-dimensional; FireAntV3’s geometry and motion simplify the problem considerably. Each dock is fixed relative to the robot’s center, limiting possible connection configurations, and its discrete flipping gait provides predictable rotation angles and step heights. These constraints transform what is typically a difficult 3D localization problem into one that can be estimated fairly accurately using only local contact sensing and proprioception.
Figure~\ref{fig:contact_constraints} illustrates how FireAntV3’s geometry and sensing capabilities progressively constrain the localization problem.

Initially, proprioceptive sensing from the IMU limits each robots orientation to a small set of candidate configurations. Identifying which previously added robot is connected can provide relational context that further narrows the feasible positions. Each FireAntV3 module consists of three spherical docks at fixed offsets from the robot’s center, and the robot can detect which of its docks is engaged with a neighbor. Recognizing specific dock-to-dock contacts imposes geometric constraints, as the distance between connected docks is fixed by the sphere diameter of the robot. Additional connections (e.g., Dock 1 of robot A to Dock 1 of robot B) further reduce the solution space. Finally, because FireAntV3 moves in discrete flips with known rotation angles and unit axes, the robot’s height ($z$) can be calculated directly from its motion history without external measurement.
 Through this layered process; combining orientation from the IMU, dock identity, and the geometric tangency and flipping history; the initially large configuration space collapses into an almost unique relative placement.

\section{Methodology}

\subsection{Target platform}

\begin{figure}
    \centering
    \includegraphics[width=\columnwidth]{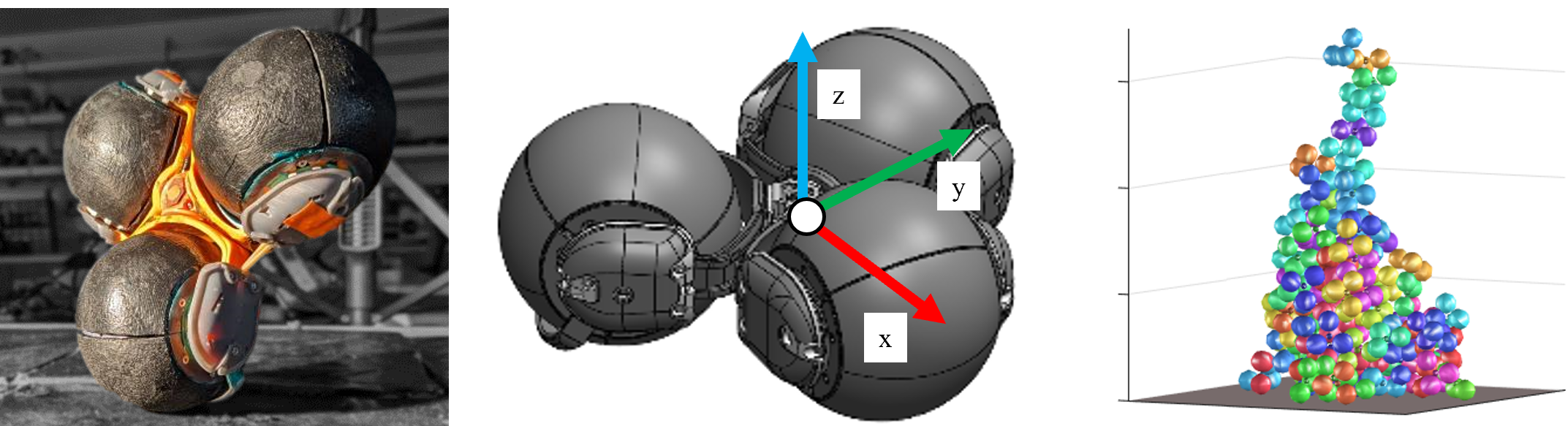}
    \caption{(Left) The FireAnt V3 robot \cite{swissler2023fireantv3}; (Center) Orientation of the axes in a simulated FireAnt robot; (Right) A tower constructed by the ReactiveBuild algorithm \cite{swissler2021reactivebuild,swissler2024behavioral}.}
    \label{fig:previous_work}
\end{figure}

Our contact-based localization approach targets the simulated models of FireAnt-style robots (shown in Fig. \ref{fig:previous_work}), specifically \textbf{FireAntV3}. FireAntV3 is the latest generation of the FireAnt series, a set of modular, self-climbing robots designed for decentralized self-assembly in contact-rich environments\cite{swissler2023fireantv3}. Each module has a \textit{three-sphere design} and employs a \textit{flipping locomotion mechanism}, enabling traversal over arbitrarily arranged peer robots. Their \textit{continuous docking surfaces} support strong, orientation-independent connections, allowing attachment at any point or angle without requiring precise alignment. FireAntV3 uses vibration-based neighbor identification and detects vibrations by measuring changes in consecutive 3D acceleration vectors with its IMUs. This allows it to distinguish peer-to-peer contacts from environmental noise and to identify which dock is engaged. The same IMU data also provides orientation, enabling FireAntV3 to localize both contact and pose.

FireAntV3 relies only on onboard IMUs for sensing, making it far simpler and cheaper than conventional localization approaches. External fiducials, motion capture systems, or GNSS are impractical for FireAnt and other self-assembly robots: line-of-sight is often blocked within dense 3D towers, GPS is unavailable indoors or underground, and lattice-constrained algorithms do not extend to the amorphous structures. 
  Beyond individual sensing, FireAntV3 has also provided the foundation for an algorithmic study of self-assembly. In particular, the \textit{ReactiveBuild} algorithm was developed and validated in simulation using FireAnt-style robots, showing how local contact and stress cues can drive the growth of amorphous, environment-adaptive structures.
 ReactiveBuild enables modular robots like FireAntV3 to self-assemble structures in 
complex environments without needing a pre-defined shape or external planning\cite{swissler2021reactivebuild}. 
Rather than building a rigid template, robots react to local conditions and
other—growing structures, such as towers, chains, cantilevers, or bridges, based on 
where stress is high or where structural support is needed. Robots decide where to 
go based on goal direction and contact information. We extend this idea of using contact information in order to effectively calculate each robots position. This turns FireAntV3's contact interaction into both construction and localization.

\subsection{Proposed Localization Method}
Our proposed contact-driven localization method builds on the sensing and interaction primitives of FireAntV3, and integrates naturally with the ReactiveBuild self-assembly framework. As new robots join a growing structure, the algorithm incrementally estimates their poses using only local contact sensing and proprioception. The first robot is designated as an anchor, and subsequent robots are localized relative to the anchor and FireAnts that have joined the structure (and have had their own positions calculated). In this way, each estimate builds upon earlier estimates. Although each FireAnts calculated location is relative to the anchor, as the number of robots increases, the given information restricts the possible configurations of robots within the structure;additionally, robots in contact with the ground provide a known $z$-height, offering a partial but independent vertical constraint that further bounds positional estimates. improving the overall accuracy. 

\subsection{Algorithm Overview}
\noindent \textbf{1) Initialization:}  
When a new robot joins the structure, it first identifies previously localized neighbors with which it has direct physical contact. The centers of these neighbors are averaged to compute an initial position estimate. To resolve overlap and maintain spatial separation, the estimate is then projected outward along the robots local $z$-axis (See Fig. \ref{fig:previous_work}) by a fixed scalar offset referred to as the \emph{scale factor}. 

\noindent \textbf{2) Localization refinement using virtual forces:}  

\paragraph*{2.a) Attraction and repulsion from neighbors}

The initial estimate is iteratively refined over $T$ steps using virtual forces derived from dock-to-dock interactions. For each contacted neighbor, the Euclidean distance $d$ and unit direction $\hat{\mathbf{d}}$ between connected dock pairs are computed. The virtual force applied to the robot’s center is given by:
\begin{equation}
\mathbf{F} =
\begin{cases}
A \cdot \alpha_t \cdot (d - D) \cdot \hat{\mathbf{d}}, & d > D, \\
- B \cdot (1 - \alpha_t) \cdot (d - D)^4 \cdot \hat{\mathbf{d}}, & d < D,
\end{cases}
\end{equation}
where $D$ is the desired dock spacing, $A$ and $B$ are attraction/repulsion gains (constants obtained by optimizing sum of squared error across all configurations), and $\alpha_t$ is a ramping factor. Early iterations prioritize repulsion to prevent overlap, followed by attraction to converge toward equilibrium spacing. This case reflects settings in which communication is limited to local neighbors.
\paragraph*{2.b) Global disconnected repulsion}
In a purely local-communication setting, a robot only knows which of its own docks are in physical contact; it has no information about nearby docks that are not contacting. The \emph{undocked-repulsion} term therefore assumes a minimal global channel: robots periodically broadcast an estimated pose proxy (e.g., center and dock position). Using these messages, any pair of docks $(d_i,d_j)$ that are not connected but lie within a collision threshold $d_{ij}<D$ exert a repulsive force to discourage crowding

\begin{equation}
\mathbf F^{\text{undock}}_{ij} =
\begin{cases}
-\,B\,(1-\alpha_t)\,\bigl(d_{ij}-D\bigr)^{4}\,\hat{\mathbf d}_{ij} & \text{if } d_{ij}<D\\[2pt]
\mathbf 0 & \text{otherwise}
\end{cases}
\end{equation}

where $d_{ij}$ is the Euclidean distance between the docks and $\hat{\mathbf d}_{ij}$ is the unit vector from dock $j$ to dock $i$. This mirrors the connected-dock repulsion but applies it to nearby \emph{unconnected} docks when the minimal global channel is available.
\label{sec:undock}

\paragraph*{2.c) Z-Axis Correction}  
Each robot maintains a discrete record of its flipping history, including the rotation angles and axes of successive moves. 
Because FireAntV3 modules are initialized from a known base height, the $z$-position of each robot can be inferred from the cumulative vertical displacement implied by its flips. 
To exploit this, we introduce an additional virtual force directed from the current estimated position toward the $z$-coordinate derived from flipping history:  
\begin{equation}
\mathbf{F}_z = k_z \bigl(z_{\text{flip}} - z_{\text{est}}\bigr),
\end{equation}
where $k_z$ is a correction gain and $z_{\text{flip}}$ denotes the height predicted from flip records. 
This correction is especially effective in tower-like assemblies, where vertical stacking dominates, as it reinforces consistent alignment along the $z$-axis across robots. 
.

\begin{algorithm}[!t]
\caption{Virtual Force-Based Localization Algorithm}
\label{alg:vf_localization}
\SetKwInOut{Input}{Input}
\SetKwInOut{Output}{Output}

\Input{Swarm structure $S$, base robot $b$ with known pose}
\Output{Estimated positions $P$ for all robots}

\ForEach{new robot $r$ joining the structure}{
    $P_r \gets \text{average}(P_{N_r}) + s\hat{z}$ \tcp*[r]{Initialization}

    \Repeat{error $<$ threshold or max iterations reached}{
        \ForEach{neighbor $n \in N_r$\tcp*[r]{Refinement}}{
            \ForEach{dock pair $(d_r, d_n)$}{
                Compute distance $d_{r,n}$ and direction $\hat{u}_{r,n}$\;
                \eIf{docks are connected}{
                    \uIf{$d_{r,n} > D$}{Apply attraction: $F \propto (d_{r,n} - D)$}
                    \uElseIf{$d_{r,n} < D$}{Apply repulsion: $F \propto -(d_{r,n}-D)^4$}
                }{
                    \If{$d_{r,n} < D$}{Apply repulsion to avoid overlap}
                }
            }
        }
        Update $P_r$ using cumulative forces\;
        
        \If{flipping history is available}{
           
            Apply attraction:\;
            \vspace{0.1cm}
            $F \propto (z_{\text{flip}} - z_{\text{est}})$\;
        }
    }
}
\Return{$P$}
\end{algorithm}
    
\section{Simulation-Based Validation}

To validate the proposed localization algorithm, we use a virtual swarm of FireAntV3 robots in the ReactiveBuild simulation environment. Robots move toward a target region by flipping around their spherical docks and attaching sequentially; each robot localizes itself relative to previously placed neighbors. The simulator waits for each attachment to complete before introducing the next robot. We consider two self-assembly scenarios: \textit{towers} (vertical stacking) and \textit{cantilevers} (horizontal extension from a base). For each configuration, we generate and analyze 
25 independent structures with  100 robots to quantify localization accuracy.
\subsection{Evaluation Metrics}

Let $\mathbf{p}_i \in \mathbb{R}^3$ denote the \emph{true} center of robot $i$ and $\hat{\mathbf{p}}_i \in \mathbb{R}^3$ its \emph{estimated} center.

\paragraph*{ SSE and RMSE}
We calculate the absolute localization error by
\begin{equation}
\mathrm{SSE} \;=\; \sum_{i=1}^{N} \bigl\| \hat{\mathbf{p}}_{i}-\mathbf{p}_{i} \bigr\|^{2},
\qquad
\end{equation}
\begin{equation}
\mathrm{RMSE} \;=\; \sqrt{\frac{1}{N}\sum_{i=1}^{N} \bigl\| \hat{\mathbf{p}}_{i}-\mathbf{p}_{i} \bigr\|^{2}},
\end{equation}
where RMSE (Root mean square error), reported in robot radii, reflects the typical per-robot localization error.

\paragraph*{Procrustes disparity (shape similarity)}
RMSE measures accuracy in the world frame, so global translation or rotation, like anchor drift, inflates error even if the relative arrangement of robots is correct. Following geometric morphometrics in biology (where shapes are compared after removing pose), we also report \emph{Procrustes shape disparity} $d$, computed after optimally aligning the estimated configuration to ground truth with a rigid transform (rotation + translation; no scaling, no reflection). We first solve the optimization problem
\begin{equation}
\min_{\mathbf{R},\,\mathbf{t}} \;\sum_{i=1}^{N} \bigl\| \mathbf{p}_{i} - \bigl(\mathbf{R}\,\hat{\mathbf{p}}_{i} + \mathbf{t}\bigr) \bigr\|^{2}
\quad \quad 
\end{equation}
for $R$ and $t$.
Let $\mathbf{z}_i=\mathbf{R}\,\hat{\mathbf{p}}_{i}+\mathbf{t}$ be the aligned estimates and $\bar{\mathbf{p}}=\tfrac{1}{N}\sum_{i=1}^{N}\mathbf{p}_i$ the true centroid. We report the \textbf{Procrustes disparity}
\begin{equation}
d \;=\; \frac{\sum_{i=1}^{N}\bigl\|\mathbf{p}_i-\mathbf{z}_i\bigr\|^{2}}{\sum_{i=1}^{N}\bigl\|\mathbf{p}_i-\bar{\mathbf{p}}\bigr\|^{2}} \;\in\;[0,1],
\end{equation}
where $d=0$ indicates identical configurations after alignment and $d=1$ indicates no geometric similarity\cite{{disparity}}.
This makes $d$ sensitive to structural fidelity (who is where relative to whom) while being invariant to global pose. As a simple example, consider two 100-robot towers with identical stacking order—one perfectly vertical, the other inclined by $10^{\circ}$. The inclined tower can show a large RMSE because every robot is displaced in the world frame, yet its shape disparity $d$ remains small once aligned, indicating that the assemblies are geometrically similar. We therefore present both metrics: RMSE for deployment-relevant absolute accuracy and $d$ to quantify how well our estimate matches the true structure.

For validation, we first run the algorithm with the full set of available features, including \emph{Global disconnected repulsion}, which applies virtual repulsive forces between nearby but unconnected docks to prevent overlap in dense regions, and \emph{Z-estimation}. We then progressively remove these features to assess their individual impact on localization accuracy.

\subsection{Tower}

To evaluate the algorithm, we simulated a scenario where the swarm constructs a vertical tower to a target height of 65 robot radii, with new robots added sequentially from random initial poses.

\subsubsection{Baseline}
\begin{figure}[t]
  \centering
  \includegraphics[width=\linewidth]{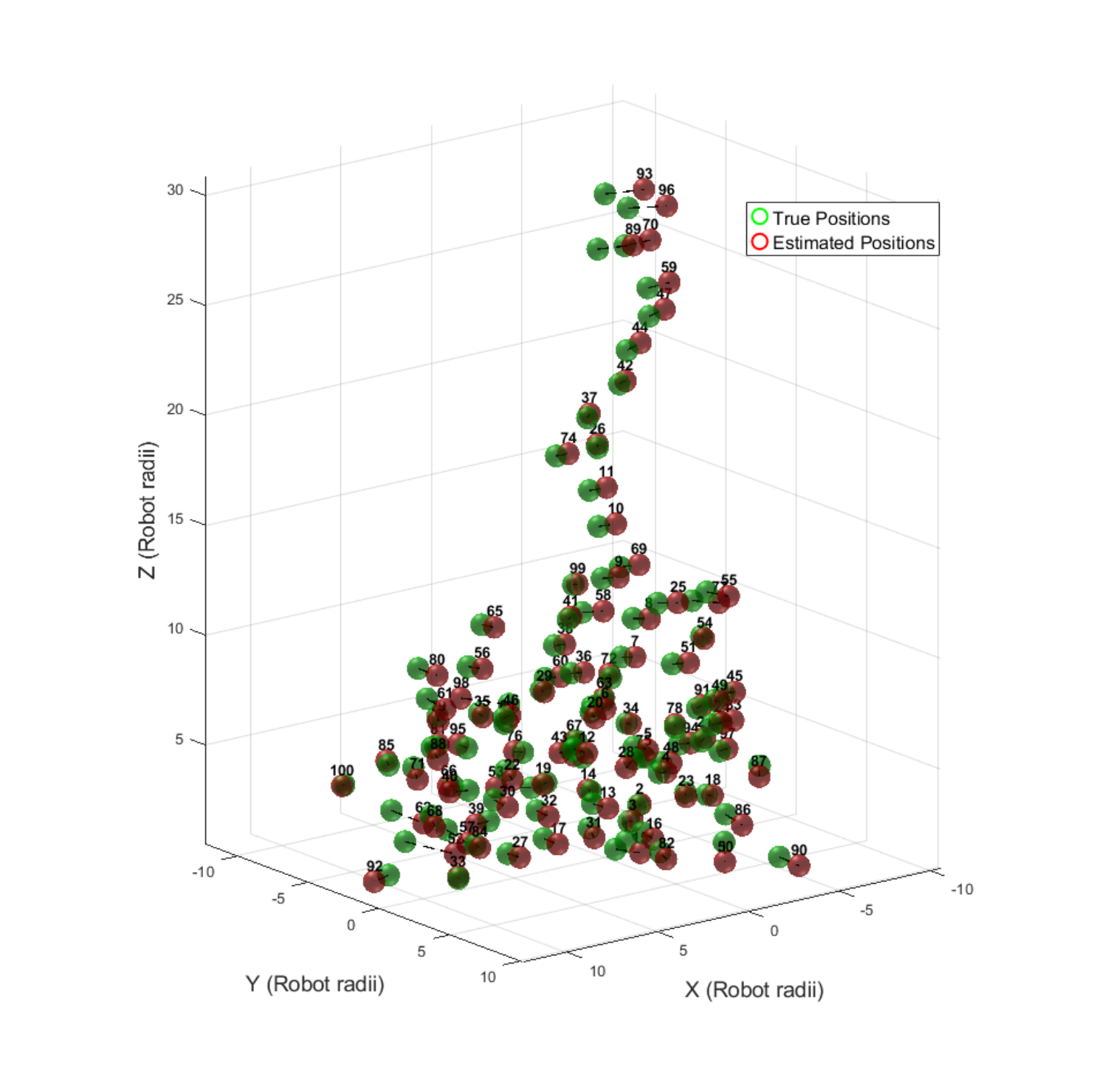}
  \caption{Estimated vs True robot positions for a tower configuration. 
  Each sphere marks a robot center.}
  \label{fig:3dplottower}
\end{figure}

\begin{figure}[t]
    \centering
    \includegraphics[width=0.99\linewidth]{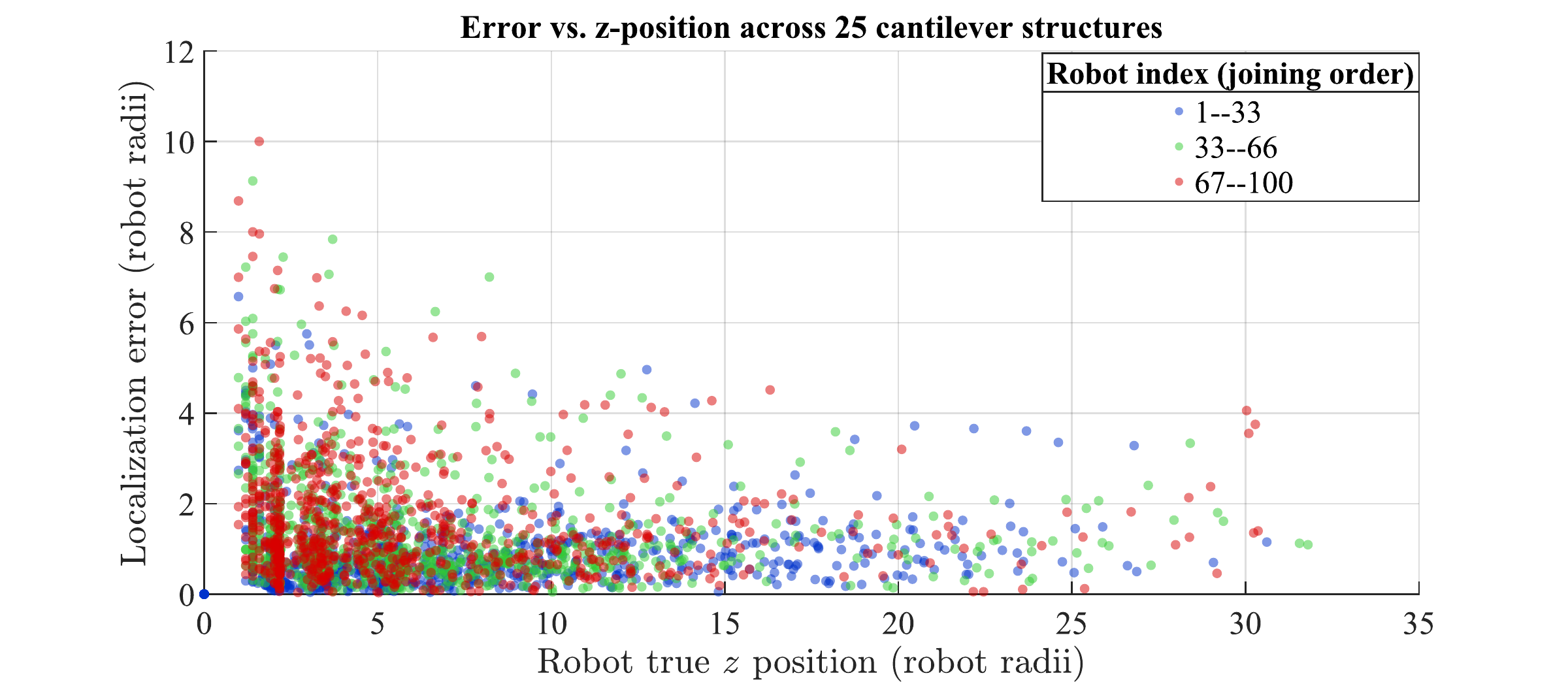}
    \caption{Per-robot localization error versus true height for 25 tower 
assemblies. As the tower grows taller, the pyramid-like geometry 
narrows the feasible region for each robot, reducing localization 
ambiguity and yielding lower error variability at greater heights. 
Robot joining order does not correlate with localization error. 
structural geometry is the dominant factor governing accuracy. }
    \label{fig:scattertower}
\end{figure}

The algorithm's performance in reducing the RMSE error is shown in Fig.~\ref{fig:rmse_fanchart_ablations} as the red curve. The initial error from the projection step is rapidly minimized, with the system converging to a final RMSE of 1.64 robot radii (approximately 54 mm). Performance was observed to be dependent on the final structure geometry, with certain tower configurations achieving an even lower RMSE of 0.7 radii (approximately 23 mm). Figure~\ref{fig:3dplottower} overlays the estimated and ground-truth robot positions for a sample configuration, demonstrating a close match.
Another important observation is that, as the tower grows, robots at greater heights exhibit lower error variation. 
This arises because their feasible placement region is progressively constrained by the hierarchical tower geometry: 
the cross-section narrows with height, reducing the number of admissible solutions. 
This trend is illustrated in Fig. \ref{fig:scattertower}, where localization error (Euclidean distance between the true position and the estimated one) decreases in variability at higher levels of the tower.

We next ablate additional sensory/interaction cues from the swarm system.

\subsubsection{Removing $Z$ information (flipping history)}
We disable the flipping-history cue used to estimate each robot’s outward $Z$ direction.
Without this orientation prior, vertical mirror ambiguities remain, which slows convergence and increases localization error (blue vs. red in Fig.~\ref{fig:rmse_fanchart_ablations}).The Procrustes disparity increases proportionally from $d=0.045$ to $d=0.051$ 
(+13\%), consistent with the +12\% increase in RMSE. This proportional degradation indicates that removing the $Z$ cue uniformly affects both absolute accuracy and structural fidelity. In other words, the error is not simply a global vertical shift of the whole assembly, but a genuine distortion of relative robot placement. Nevertheless, the moderate magnitude of both increases suggests that the algorithm degrades gracefully without the flipping-history cue, preserving the overall tower shape reasonably well.

\subsubsection{Removing global disconnected repulsion}
\begin{figure}[t]
  \centering
  \includegraphics[width=\linewidth,height=5.5cm]{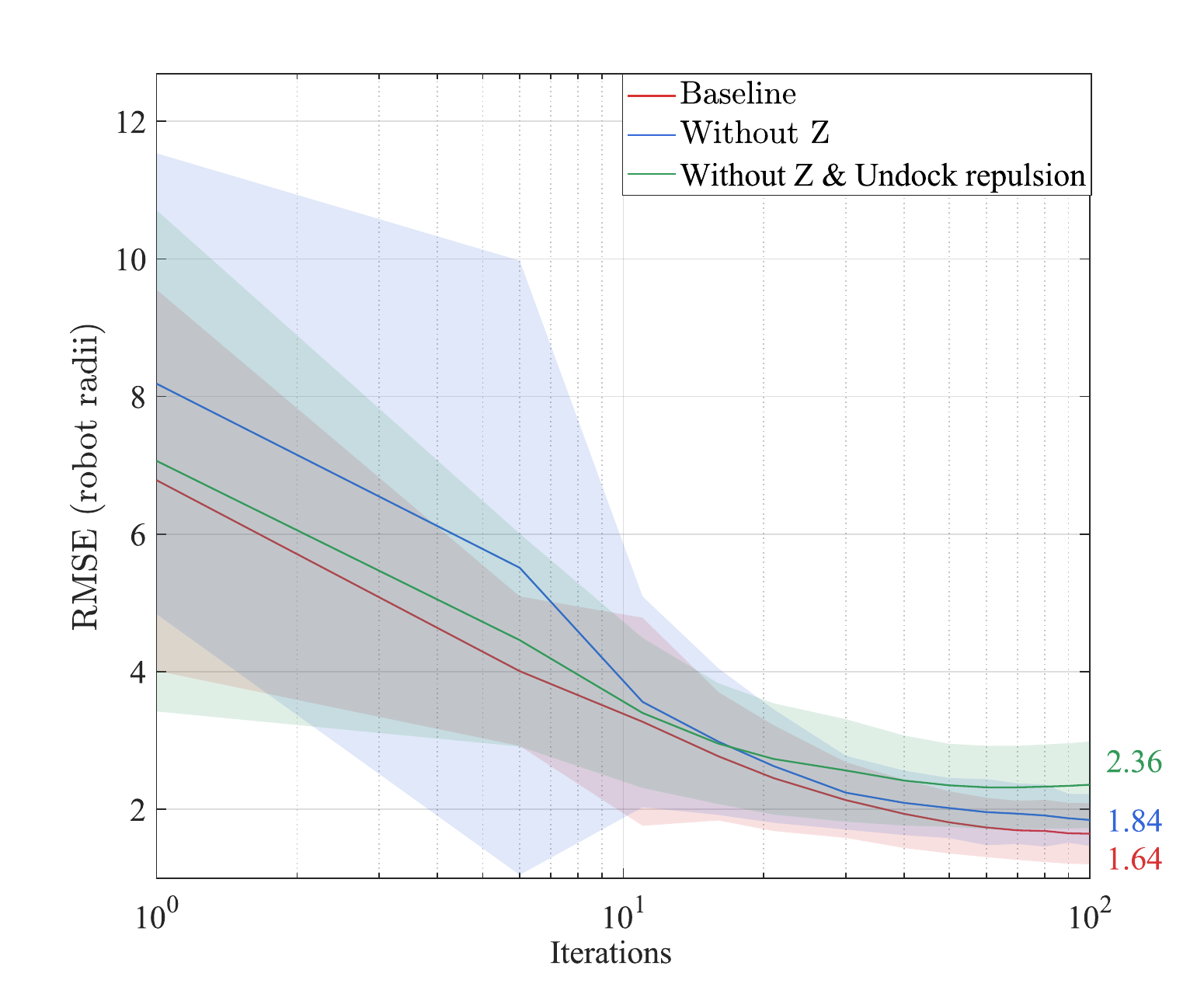}
  \caption{Convergence of localization RMSE versus iterations for the tower configuration.
  Curves show the mean across 25 runs and shaded bands denote $\pm 1$\,STD.
  {Baseline} uses $Z$ estimation from flipping history and global disconnected repulsion;
{Without $Z$} removes the flipping-history cue;
{Without $Z$ \& undock repulsion} removes both cues.
  Ablations slow convergence and increase steady-state error
  (final mean RMSE = 1.64, 1.84, and 2.36 robot radii, respectively).}
  \label{fig:rmse_fanchart_ablations}
\end{figure}

We further simplify the algorithm by removing the undock–repulsion term.
This allows non-contacting docks to collapse toward each other, biasing 
pose estimates, and yielding the largest error among conditions (green 
in Fig.~\ref{fig:rmse_fanchart_ablations}). The Procrustes disparity 
increases from $d=0.045$ to $d=0.066$ (+47\%), proportional to the 
+44\% rise in RMSE. The magnitude of degradation is substantially 
larger than the $Z$-ablation case, indicating that global disconnected 
repulsion is more critical for maintaining structural fidelity in towers. 
Without it, nearby unconnected docks collapse inward, introducing 
genuine misplacement of robots relative to their true neighbors.

\subsection{Cantilever}
\begin{figure}[t]
  \centering
  \includegraphics[width=\linewidth]{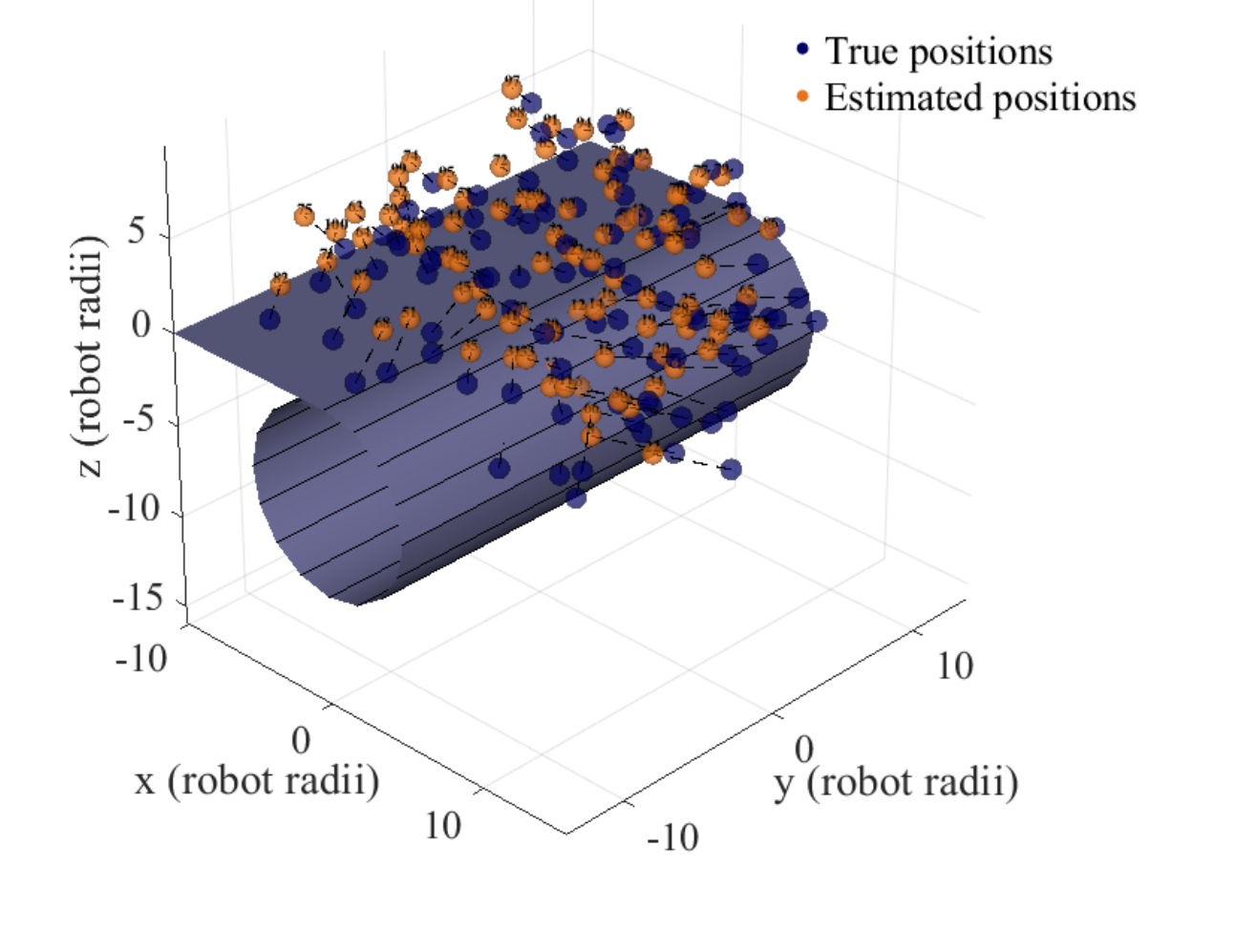}
  \caption{Estimated vs.\ true robot positions for a cantilever configuration. 
  Each sphere marks a robot center.}
  \label{fig:3dplotcantilever}
\end{figure}

  A cantilever is a structure that projects horizontally. To evaluate ReactiveBuild’s ability to produce such a form, we simulate robots growing a cantilever toward a goal located 45 units horizontally outward from the platform edge and 10 units(robot radii) vertically above it. As illustrated in Fig.~\ref{fig:3dplotcantilever}, the horizontal plane terminates at a cylindrical boundary at its edge. Robots are introduced one at a time behind the current leading module, with randomly sampled initial offsets and orientations relative to the furthest placed robot, and then attach sequentially to extend the cantilever.
\subsubsection{Baseline}
\begin{figure}[t]
    \centering
    \includegraphics[width=0.99\linewidth]{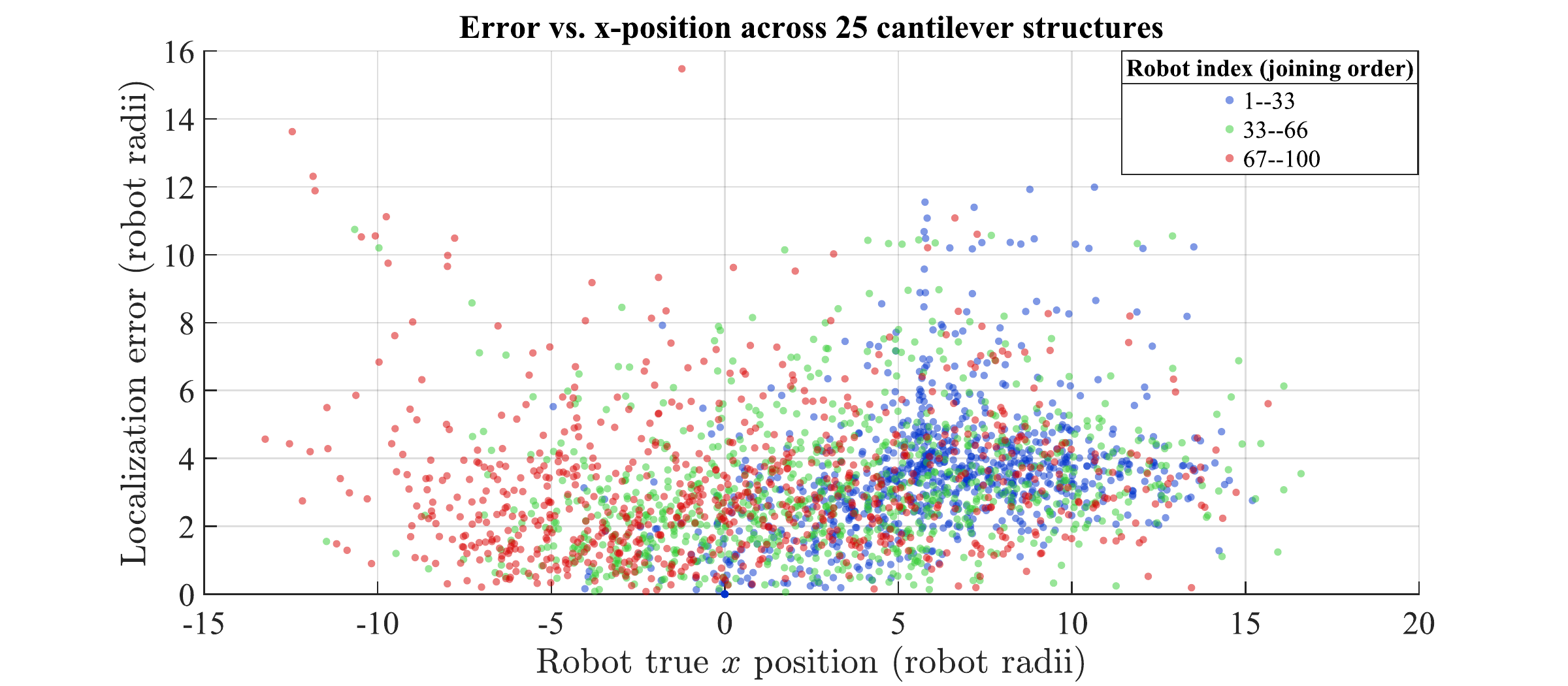}
    \caption{Per-robot localization error versus true longitude for 25 Cantilever assemblies.}
    \label{fig:scattercantilever}
\end{figure}

The red curve in Fig.~\ref{fig:convergence_rms_cantilever} is the
\emph{baseline} (all sensory cues enabled: flipping–history $z$ cue and
undocked–repulsion). Its steady–state error is higher than the tower case
(Fig.~\ref{fig:rmse_fanchart_ablations}) for three reasons. First, the cantilever grows horizontally and does not develop the pyramid-like narrowing of a tower, so the solution space does not shrink as much based on height, and the z derived from flipping history is less informative. 
Secondly, robots are localized relative to previously estimated neighbors. In towers, robots 2 and 3 attach immediately to the anchor ($N{=}1$), creating a rigid base and limiting error propagation. 
In cantilevers, the first robot that actually contacts the anchor
often appears much later in the join order (e.g., $N\!\approx\!13$), so robots
with indices 2–12 are placed relative to an unanchored chain, and their errors
accumulate. This contrast is also visible in the scatter plots: in towers (Fig.~\ref{fig:scattertower}, early-joining robots (dark blue, index $\leq$ 10) cluster among the lowest-error points, whereas in cantilevers (Fig.~\ref{fig:scattercantilever}) the dark-blue points are not concentrated at low-error regions.

Despite these challenges, the baseline converges quickly from the initial projection step and reaches a final RMSE of $3.91$ robot radii ($\approx\!92$\,mm),
i.e., $<10$\,cm—acceptable for our module dimensions.

\subsubsection{Removing the $Z$ cue (flipping history)}
In cantilevers, the vertical cue is weaker than in towers, but it is still helpful: it fixes the sign of vertical drift and reduces pose ambiguity near
the plate. Removing it increases both error and uncertainty, slowing early
convergence and raising the steady–state RMSE from \(\,3.91\,\) to
\(\,5.26\,\) robot radii (Fig.~\ref{fig:convergence_rms_cantilever}).
The wider shaded band reflects greater ambiguity when the structure grows
horizontally without the flipping-history constraint.
The Procrustes disparity increases from $d=0.166$ to $d=0.219$ (+32\%), proportional to the +35\% rise in RMSE. This consistency indicates that removing the $Z$ cue degrades both absolute accuracy and structural 
fidelity uniformly, rather than introducing a simple global offset.

\subsubsection{Removing global disconnected repulsion (in addition to flipping history)}
\begin{figure}[t]
  \centering
  \includegraphics[width=\linewidth, height=4.7cm]{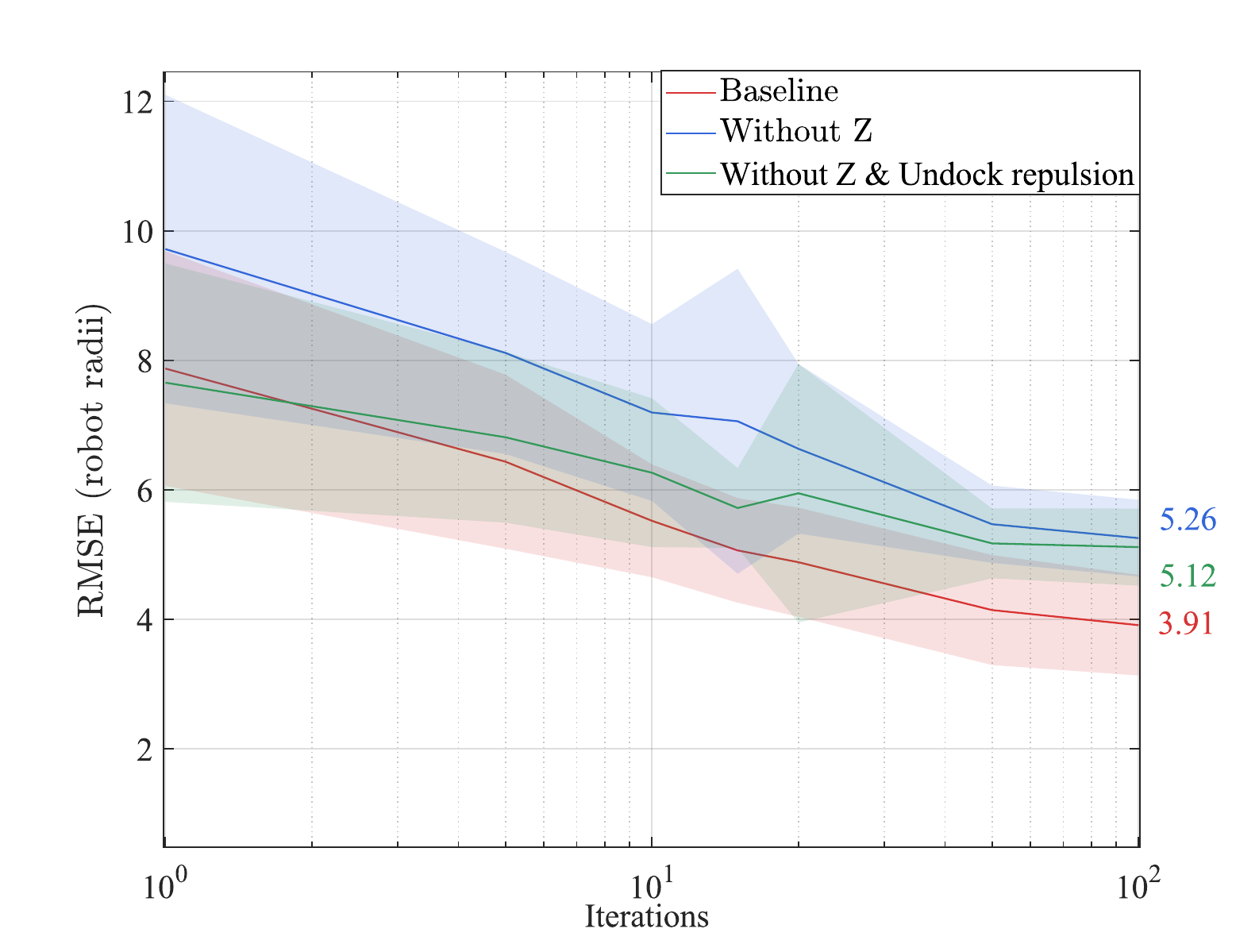}
  \caption{Convergence of localization RMSE versus iterations for the cantilever configuration.
  Curves show the mean across 25 runs and shaded bands denote $\pm 1$\,STD.
  {Baseline} uses the flipping-history $z$ cue and Global disconnected repulsion;
  {Without $Z$} removes the flipping-history cue;
  {Without $Z$ \& undock repulsion} removes both cues.
  The baseline converges to a final RMSE of $3.91$ robot radii ($\approx 92$\,mm).}
  \label{fig:convergence_rms_cantilever}
  \end{figure}

Turning off global disconnected repulsion on top of the 
no-$Z$ condition changes performance only in terms of 
final RMSE = 5.12 vs.\ 5.26 radii, and the shaded bands 
largely overlap in Fig.~\ref{fig:convergence_rms_cantilever}. 
This small effect is expected for cantilevers: robots extend 
into free space, so near but unconnected docks are rare, 
and spacing is already constrained by connected contacts 
and the plate geometry. However, the Procrustes disparity increases 
from $d=0.219$ to $d=0.251$ (+14\%), indicating that the relative arrangement of robots within the structure worsens.

\section{Discussion}
Table~\ref{tab:rmse_procrustes} summarizes RMSE and Procrustes disparity across all conditions for both tower and cantilever.
RMSE values are expressed in Robot radii units; the baseline tower accuracy of 1.64 radii corresponds to 
approximately 54\, mm error, while cantilevers reach about 92\,mm. In both cases, 
robots localize themselves with accuracy on the order of one dock diameter, which is 
sufficient for reliable attachment in self-assembly.
Localization quality depends strongly on structure geometry and join order. In towers, robots 2 and 3 attach directly to the anchor, forming a rigid base that locks pose early; errors stay low and do not accumulate (baseline RMSE $=1.64$ radii; some runs near $0.7$). Cantilevers, by contrast, exhibit a large join-index gap before the anchor’s \emph{first} physical neighbor appears (e.g., the 13th robot), so early placements drift and error accumulates (baseline RMSE $=3.91$ radii).
The global communication that enables undocked repulsion is clearly beneficial for \emph{towers} since it reduces both RMSE and shape disparity. The reason is that towers are compact, so there are many near-but-unconnected docks, and the repulsion term helps prevent local collapse and bias. For \emph{cantilevers}, the story is different: robots extend into free space, so undocked pairs are rarer; the term can lower $d$ slightly but has little impact on RMSE. The $Z$ cue (from flipping history) is a robust, low-cost constraint and should be enabled by default for both configurations.

\begin{table}[t]
\centering
\caption{RMSE and Procrustes Error across Conditions}
\resizebox{\columnwidth}{!}{%
\begin{tabular}{llcc}
\toprule
\textbf{} & \textbf{Condition} & \textbf{RMSE} & \textbf{Procrustes disparity} \\
\midrule
Tower      & Baseline              & 1.64 & 0.045  \\
           & No Z Estimation       & 1.84 (+12\%) & 0.051  (+13\%) \\
           & No global comm. or Z est. & 2.36 (+44\%)  & 0.066 (+47\%) \\
\midrule
Cantilever & Baseline              &  3.91  & 0.166 \\
           & No Z Estimation       &   5.26 (+35\%)  & 0.219 (+32\%) \\
           & No global comm. or Z est. & 5.12 (+31\%) & 0.251  (+51\%)\\
\bottomrule
\end{tabular}}
\label{tab:rmse_procrustes}
\end{table}

\section{CONCLUSIONS}
Through simulation with FireAntV3 modules, we demonstrated a contact-driven, infrastructure-free localization method that operates during freeform, environment-adaptive self-assembly. Because it relies primarily on IMU-based proprioception, which is a common capability of many self-assembling platforms, the method is not specific to FireAntV3 and can transfer with minimal changes to other modular systems. 
Motivated by the incremental nature of estimation and the join-index gap observed in cantilever assemblies, we propose exploring a local neighbor-to-neighbor \emph{back-propagation} pass as part of future work to overcome error accumulation. 
In this approach, a back-propagation step would be triggered after every ten joins to revisit recent placements, 
allowing error corrections to propagate backward. This direction of investigation could help identify 
when and how frequently back-propagation should be applied to improve overall localization.

We also plan to reduce reliance on magnetometer yaw, which can be unreliable in practice due to disturbances and vibration, by introducing a virtual torque that aligns yaw using dock geometry and neighbor constraints, paired with IMU-centric filtering. Finally, we will deploy this decentralized localization on Simlified version of Fireant2d  hardware to evaluate this method in experiments.

% \addtolength{\textheight}{-12cm}   %

\section*{ACKNOWLEDGMENT}

The authors are grateful to Roberto Torres for his 
help with the literature review.

\balance

\end{document}